\documentclass[lettersize,journal]{IEEEtran}
\usepackage{amsmath,amsfonts}
\usepackage{algorithmic}
\usepackage{algorithm}
\usepackage{array}
\usepackage[caption=false,font=normalsize,labelfont=sf,textfont=sf]{subfig}
\usepackage{textcomp}
\usepackage{stfloats}
\usepackage{url}
\usepackage{verbatim}
\usepackage{graphicx}
\usepackage{cite}
\usepackage[colorlinks=true, allcolors=blue]{hyperref}

\begin{document}
\title{Synthetic-to-real Composite Semantic Segmentation in Additive Manufacturing}

\author{Aliaksei Petsiuk, Harnoor Singh, Himanshu Dadhwal, Joshua M. Pearce

\thanks{Manuscript created October, 2022; This work was developed by Free Appropriate Sustainability Technology (FAST) research group at University of Western Ontario, Canada. This work is distributed under the GNU General Public License (GPL) 3.0 (https://www.gnu.org/licenses/gpl-3.0.en.html).}
\thanks{A. Petsiuk, H. Singh, H. Dadhwal, and J.M. Pearce are with ECE, Western University, London, ON, Canada, (e-mails: apetsiuk@uwo.ca, hsing334@uwo.ca, hdadhwal@uwo.ca, *joshua.pearce@uwo.ca).}}


 
\maketitle

\begin{abstract}
The application of computer vision and machine learning methods in the field of additive manufacturing (AM) for semantic segmentation of the structural elements of 3-D printed products will improve real-time failure analysis systems and can potentially reduce the number of defects by enabling in situ corrections. This work demonstrates the possibilities of using physics-based rendering for labeled image dataset generation, as well as image-to-image translation capabilities to improve the accuracy of real image segmentation for AM systems. Multi-class semantic segmentation experiments were carried out based on the U-Net model and cycle generative adversarial network. The test results demonstrated the capacity of detecting such structural elements of 3-D printed parts as a top layer, infill, shell, and support. A basis for further segmentation system enhancement by utilizing image-to-image style transfer and domain adaptation technologies was also developed. The results indicate that using style transfer as a precursor to domain adaptation can significantly improve real 3-D printing image segmentation in situations where a model trained on synthetic data is the only tool available. The mean intersection over union (mIoU) scores for synthetic test datasets included 94.90\% for the entire 3-D printed part, 73.33\% for the top layer, 78.93\% for the infill, 55.31\% for the shell and 69.45\% for supports.
\end{abstract}


\section{Introduction}
\IEEEPARstart{W}{ith} the current exponential growth, the amount of plastic waste could reach 250 billion tons by 2050 \cite{ref1}, vast quantities of which end up polluting the natural environment on land and in the ocean \cite{ref2}. Distributed manufacturing using additive manufacturing (AM) is reforming global value chains as it increases rapidly \cite{ref3} because there are millions of free 3-D printable consumer product designs and 3-D printing them results in substantial cost savings compared to conventionally-manufactured commercial products \cite{ref4,ref5}.

The growing popularity of 3-D printing is playing a notable role in the problem of recycling as 3-D printed products rarely have recycling symbols \cite{ref6}, use uncommon polymers \cite{ref7}, and are increasing the overall market of plastic materials \cite{ref8}. This is not only caused by additional plastic products, but also from disturbing failure rates. Inexperienced 3-D printer users were estimated to have failure rates of 20\% \cite{ref9}. Even experienced professionals working in 3-D print farms, however, have failure rates of at least 2\% \cite{ref10}. The probability of a manufacturing defect increases with the size and print time of the object (e.g. using large scale fused filament printers \cite{ref11} or products \cite{ref12,ref13} or fused granule printers \cite{ref14,ref15}), which can magnify the materials waste created from even a small percentage of failures. It is clear that the ability to automatically detect deviations in AM will significantly help to reduce material waste and the time spent on reproducing failed prints.

As recent studies \cite{ref16} show, computer vision is the dominant tool in analyzing AM and extrusion-based 3-D printing processes. For example, Ceruti et al. \cite{ref17} utilized data from computer-aided design (CAD) files that are used in the first step of the design of a 3-D printed component. Then further down the software toolchain, Nuchitprasitchai et al. \cite{ref18}, Johnson et al. \cite{ref19}, and Hurd \cite{ref20} developed failure analysis based on comparison with the Standard Tessellation Language (STL) files used at the slicing step in most 3-D printing. Further still, both Jeong et al. \cite{ref21} and Wasserfall et al. \cite{ref22} used instead the G-code files that provide the 3-D printer with spatial toolpath instructions for printing parts. The 3-D printing software toolchain does not need to be used at all as several approaches use comparisons with reference data \cite{ref23,ref24} or ideal 3-D printing processes \cite{ref25,ref26}. In addition, a 3-D reconstruction-based scanning method for real-time monitoring of AM processes is also possible \cite{ref27}. In previous works, the authors considered the possibilities of detecting critical manufacturing errors using classical image processing methods \cite{ref28}, as well as employing synthetic reference images rendered with a physics-based graphics engine \cite{ref29}.
\IEEEpubidadjcol

The popular open source Spaghetti Detective application \cite{ref30,ref31} is also a direct confirmation of the effectiveness of visual monitoring. An analysis of the Spaghetti Detective's \cite{ref30} user performance database collected over 2.3 years showed that 24\% of all 5.6 million print jobs were canceled, which can be represented as wasted 456 hours of continuous printing compared to 5,232 hours of printing where all the print jobs were finished (Figure 1).

\begin{figure}[!h]
\centering
\includegraphics[width=3.4in]{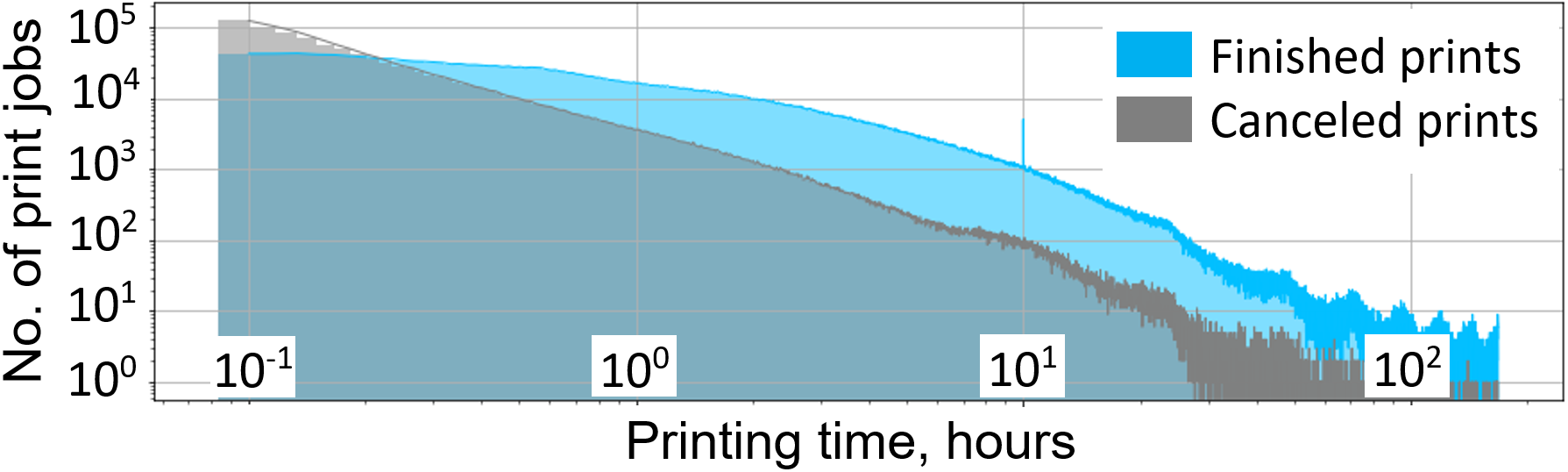}
\caption{Analysis of 3-D printer users' activity for 2.3 years. The runtime distribution shows a 24\% failure rate for all 5.6 million printing tasks longer than 5 minutes.}
\label{fig_1}
\end{figure}

This statistics, however, does not include over a million canceled print jobs less than 5 minutes long, which are assumed to be the initial bed-leveling issues and cannot, therefore, indicate manufacturing failures. It also does not consider the working time of human operators spent on starting later canceled printing tasks.

Semantic segmentation \cite{ref32} of both the entire manufactured part and its separate structural regions at the stage of production of each layer will expand the capacity of visual analysis of AM processes and will allow getting closer to the detection and localization of individual production errors. This will allow printing defects to be corrected in situ, where each successive layer can be modified depending on the deviations found in the previous stage, thus improving both the mechanical and aesthetic performance of the entire object. It also significantly reduces the requirements for camera positioning accuracy and calibration, eliminating the need for visual markers and rigid holders.

In the previous work \cite{ref29}, the authors demonstrated the ability of Blender \cite{ref33}, a free and open-source physics rendering engine, to generate photorealistic images of ideal 3-D printing processes based on existing G-code files. This work served as a milestone in the development of a deep-learning-based approach presented in this paper for the semantic segmentation of structural elements in 3-D printing environments.

Using a synthetic dataset, however, comes at a cost of a domain shift, which is often strongly associated with appearance changes \cite{ref34}. When the source (synthetic images) and target (real images) domains are semantically related, but are different in visual representation, direct propagation of learned knowledge about one domain to another can adversely affect segmentation performance in the latter domain. Therefore, Domain Adaptation (DA) is needed to learn generalized segmentation rules in the presence of a gap between the source and target dataset distributions \cite{ref34,ref35}.

There are examples in the literature of successful synthetic-to-real (sim-to-real) DA applications. Imbusch, Schwarz, and Behnke \cite{ref36} proposed an unsupervised Generative Adversarial Network (GAN)-based DA approach to a robotics environment image dataset that provides performance close to training on real data and does not require annotations of real robotic setups. Li et al. \cite{ref37} presented a semantically aware GAN-based neural network model for virtual-to-real urban scene adaptation with the ability to store important semantic information. Lee et al. \cite{ref38} introduced a sim-to-real vehicle identification technique consisting of DA and semi-supervised learning methods.

Domain adaptation, however, is a separate area of research and is not covered in this article. The possibility of applying a cycle-consistent adversarial network (CycleGAN) \cite{ref39}, an image-to-image translation method, was considered for segmentation improvement, as generative adversarial networks can perform a significant role in domain adaptation techniques and be used in future research. This work, therefore, combines the following contributions:

\begin{itemize}
\item{a technique for generating synthetic image-mask pairs of layer-by-layer ideal 3-D printing processes has been developed for subsequent neural network training;}
\item{three independent labeled synthetic image datasets for (a) entire part, (b) top layer, (c) infill, shell, and supports for 3-D printed objects have been created;}
\item{a neural network was trained for semantic segmentation of the entire printed part, as well as its current printing top layer and internal structure;}
\item{image-to-image translation approach to improve segmentation results have been explored.}
\end{itemize}

All the above steps are sequentially described in this article after first reviewing related works in detail. The results will discuss the potential for localizing of 3-D printed parts in the image frame and applying image processing methods to its structural elements for subsequent detection of manufacturing deviations.

\section{Background}
Semantic image segmentation problems represent an actively developing area of research in deep machine learning \cite{ref32,ref40}. The main limiting factor, however, is the difficulty of obtaining annotated databases for training machine learning architectures. This approach requires thousands of images with labeled masked regions, which is an extremely difficult and time-consuming task: manual annotation of a single image with pixel-by-pixel semantic labels can take more than 1.5 hours \cite{ref41}.

The use of synthetic images, in turn, allows getting a segmented training database conditionally "free of charge", since masked regions of interest can be automatically annotated when creating virtual physics-based renders. In addition, advances in computer graphics make it possible to generate an almost unlimited amount of labeled data by varying environmental parameters in ranges that are difficult to obtain in real conditions \cite{ref34}. The success of simulated labeled data is clearly illustrated in already classic GTA5 \cite{ref42} and Synthia \cite{ref43} image sets.

There are many examples of applying synthetic datasets to solve real-world practical problems. Nikolenko \cite{ref44} presented an up-to-date technological slice of the use of synthetic data in a wide variety of deep learning tasks. Melo et al. \cite{ref45} outlined the most promising approaches to integrating synthesized data into deep learning pipelines. Ward, Moghadam, and Hudson \cite{ref46} used a real plant leaves dataset augmented with rendered images for instance leaves segmentation to measure complex phenotypic traits in agricultural sustainability problems. Boikov et al. \cite{ref47} presented a methodology for steel defects recognition in automated production control systems based on synthesized image data.

Several researchers introduced artificial intelligence (AI)-based methods into AM field to classify the quality of manufacturing regions, as well as to segment failed areas in 3-D printing processes. Valizadeh and Wolff \cite{ref48} provided a comprehensive overview of neural network applications to several aspects of the AM processes. Banadaki et al. \cite{ref49} proposed a convolutional neural network (CNN)-based automated system for assessing surface quality and internal defects in AM processes. The model is trained on captured images during material layering at various speeds and temperatures and demonstrates 94\% accuracy in five failure gradations in real time. Saluja et al. \cite{ref50} utilized deep learning algorithms to develop a warping detection system. Their method extracts the layered corners of printed components and identifies warpage with 99.3\% accuracy. Jin et al. \cite{ref51} presented a novel CNN-based method incorporating strain to measure and predict four levels of delamination conditions. These works, however, solve the specific sets of certain manufacturing problems and do not allow scaling and generalization of the developed algorithms. 

Analysis based on semantic segmentation, in turn, has significant potential for detecting and evaluating a wide range of manufacturing defects. Wong et al. \cite{ref52,ref53} have demonstrated U-Net CNN 3-D volumetric segmentation in AM using medical imaging techniques to automatically detect defects in X-ray computed tomography images of specimens with a mean intersection over union (mIoU) value of 99.3\%. Cannizzaro et al. \cite{ref54} proposed an AI in-situ emerging defects monitoring system utilizing the automatic GAN-based synthetic image generation to augment the training data set. These functions are built into a holistic distributed AM platform that allows storing and integrating data at all manufacturing stages. Davtalab et al. \cite{ref55} presented a neural network automated system of semantic pixel-wise segmentation based on one million images to detect defects in 3-D printed layers.

Having an open structured annotated database for additive manufacturing will create considerable opportunities for the development of failure detection systems in the future. Segmentation and localization of individual structural elements of manufacturing objects can make it easier to detect and track erroneous regions when they occur.

\section{Methods}
Based on the most common words in 3-D print filenames stored in the Spaghetti Detective database \cite{ref30}, sets of labeled images of printed products at various stages of their production were generated in the physics-based graphics engine \cite{ref29}. These image-mask pairs were further used to train neural networks for the tasks of visual segmentation of manufactured parts and their structural elements. Additionally, the possibilities of image-to-image style translation were also explored to reduce the domain gap and increase the segmentation precision. The segmentation efficiency was tested both on synthetic renders outside of training sets and on real images. Data and source code for this project can be obtained from the Open Science Framework (OSF) repository \cite{ref56}.

\subsection{Creation of synthetic image datasets}
\subsubsection{Selecting CAD designs for rendering}
More than 5.6 million filenames were partitioned into meaningful lexical parts and processed in Spaghetti Detective's user performance database \cite{ref30} analysis to create a dictionary of the most frequently used words (Figure 2). These print jobs were performed by 49,000 unique users on 57,000 different 3-D printers. The average print time was 3.6 hours.

\begin{figure}[!h]
\centering
\includegraphics[width=3.4in]{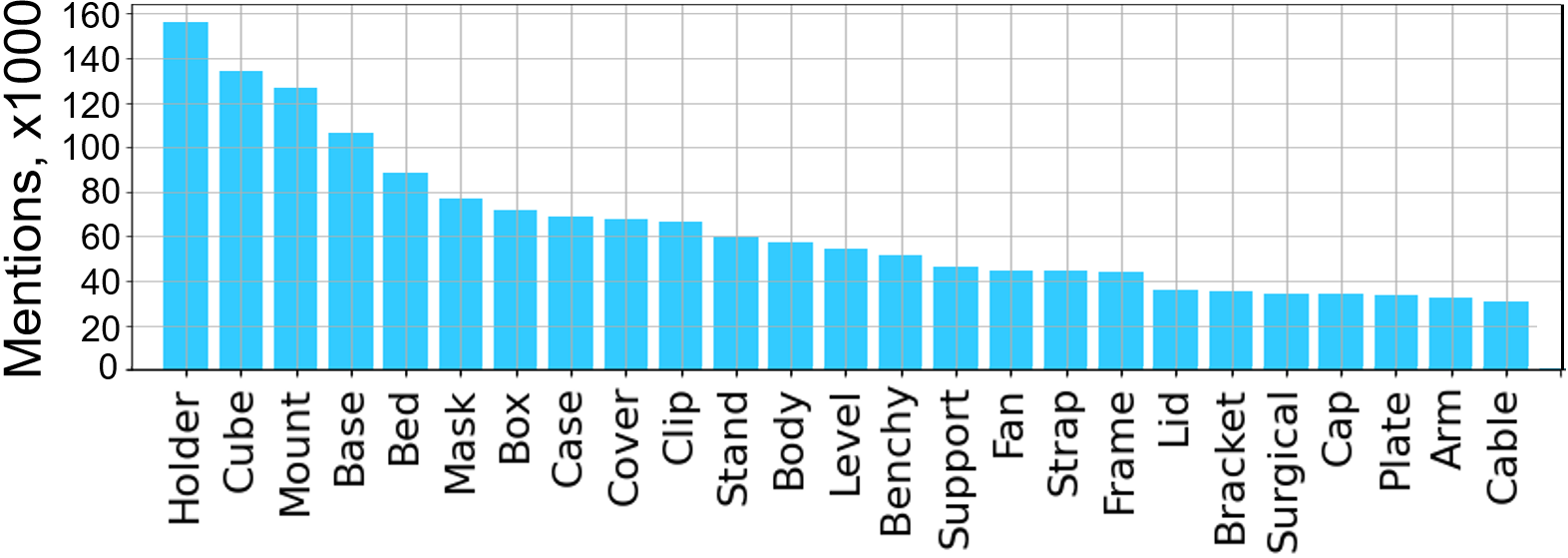}
\caption{Distribution of the 25 most frequently used words in file names for 3-D printing. A detailed analysis of the users' print tasks database is given in the source file repository \cite{ref56}.}
\label{fig_2}
\end{figure}

Based on the compiled dictionary, a set of random Standard Tessellation Language (STL) files was collected from Thingiverse \cite{ref57}, an open catalog of widely used computer-aided designs (CADs) for 3-D printing, for further processing. These files formed the basis for generating a database of synthetic images. A complete list of used CAD designs is in the OSF repository \cite{ref56}.

\subsubsection{Graphics rendering pipeline}
All the selected STL files were converted into G-codes in free MatterControl software \cite{ref58} maintaining the same slicing parameters: 0.3 mm layer height, 0.4 mm nozzle diameter, 4 perimeters, and 30\% grid infill. The resulting G-codes were further parsed layer by layer in the Blender \cite{ref33} programming interface, where the extruder trajectory is converted into a set of curves with a controllable thickness parameter and preset material settings. Each G-code layer is thus transformed into an independent 3-D object. The whole rendering process is illustrated in the following diagram (Figure 3).
 
\begin{figure}[!h]
\centering
\includegraphics[width=3.4in]{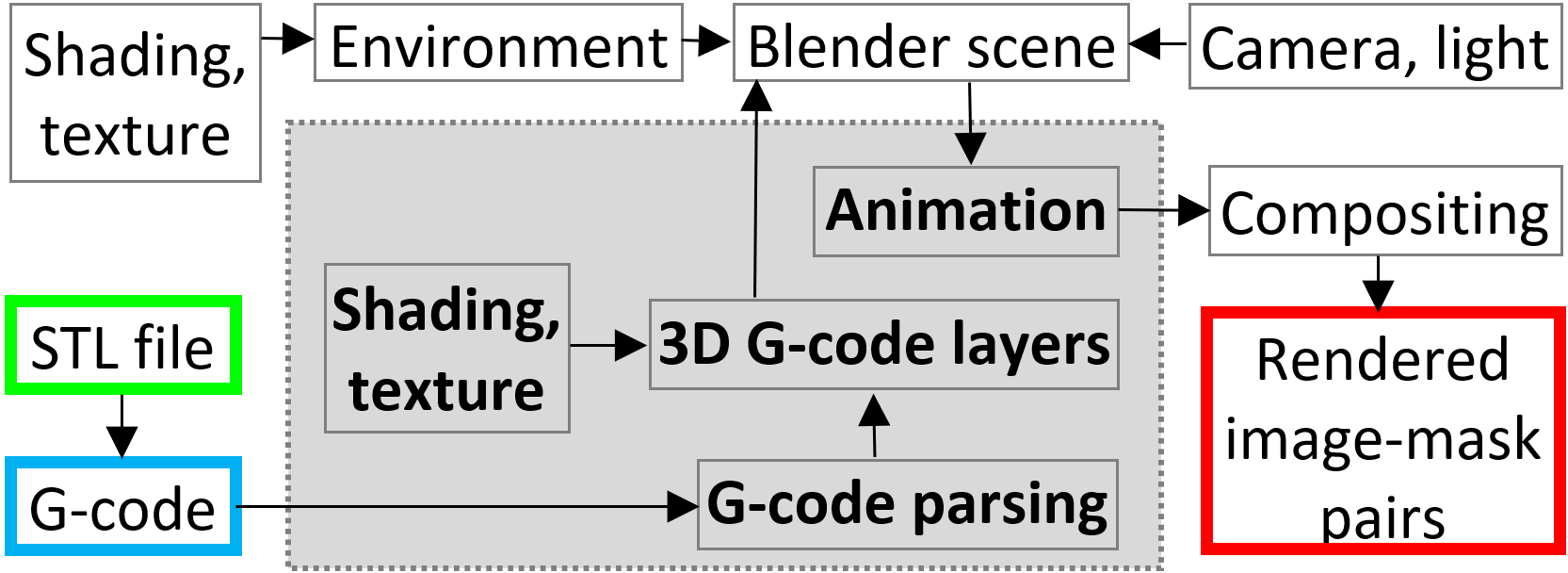}
\caption{Synthetic AM database creation pipeline. Each 3-D part in the form of an STL (green) file is converted into a set of printer tool head trajectories (G-code, blue), which is the input parameter of the automated scripted section (gray). Blender environment (textures, camera, lights) and compositing settings can also be automated in the future. The image-mask pairs (red) are the result of a frame-by-frame animation rendering for each individual G-code file.}
\label{fig_3}
\end{figure}

The functional component of the repository \cite{ref59} was used as a basis for importing G-code files into the graphics engine. To create photorealistic renders, scenes similar to real physical environments were created in Blender. The position of the camera, as well as the degree of illumination and the location of light sources, were chosen to closely match the actual workspace. (Figure 4).
 	 
\begin{figure}[!h]
\centering
\includegraphics[width=3.4in]{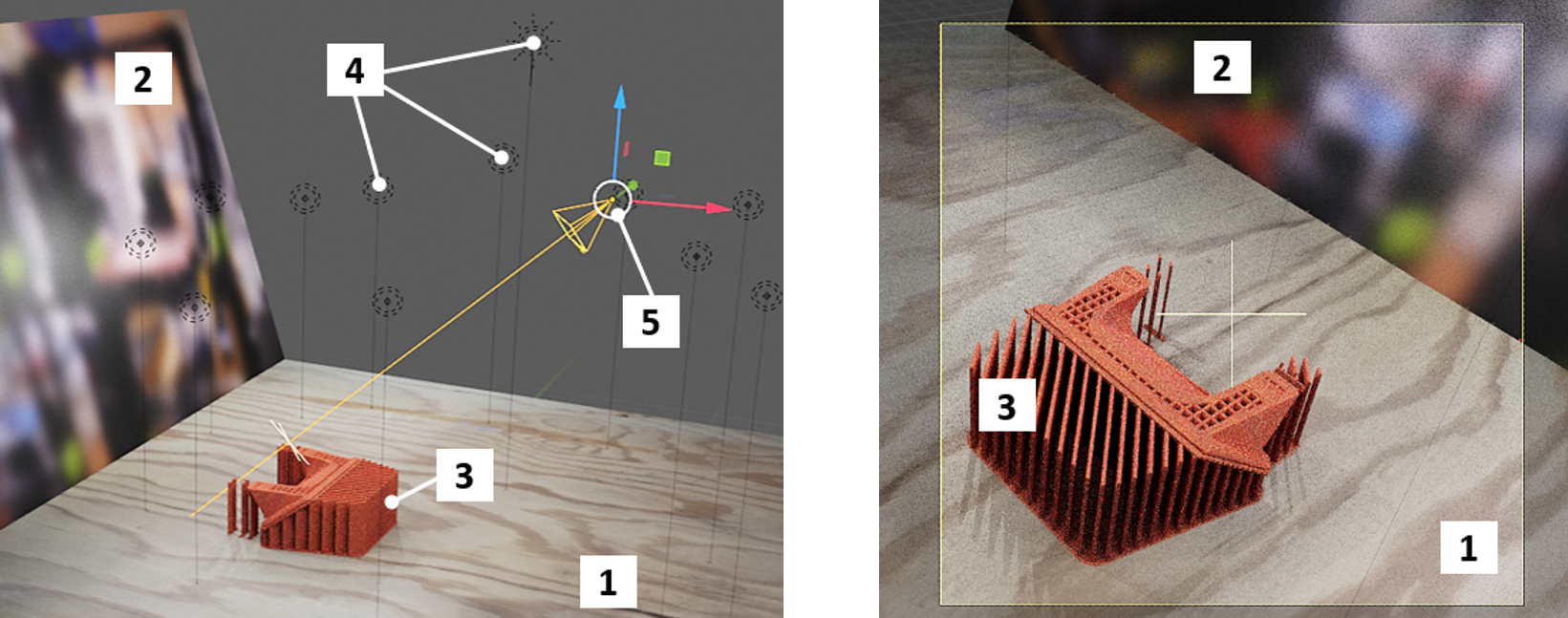}
\caption{Blender Scene: user window (left) and the virtual camera viewport (right). 1--–printing bed texture, 2---background image plane simulating ambient environment, 3–--rendered manufactured part, 4–--light sources with variable locations, 5–--camera with variable location.}
\label{fig_4}
\end{figure}

The whole scene, in addition to the printed part, includes components such as point light sources to create diverse heterogeneous all-round illumination, the "Sun" to create uniform background lighting, a flat printing surface with realistic texture and reflectivity, and a plane with a superimposed blurred background image to create an illusion of defocused ambient environment.

Figure 5 illustrates several examples of realistic textures for the printing bed plane.

\begin{figure}[!h]
\centering
\includegraphics[width=3.4in]{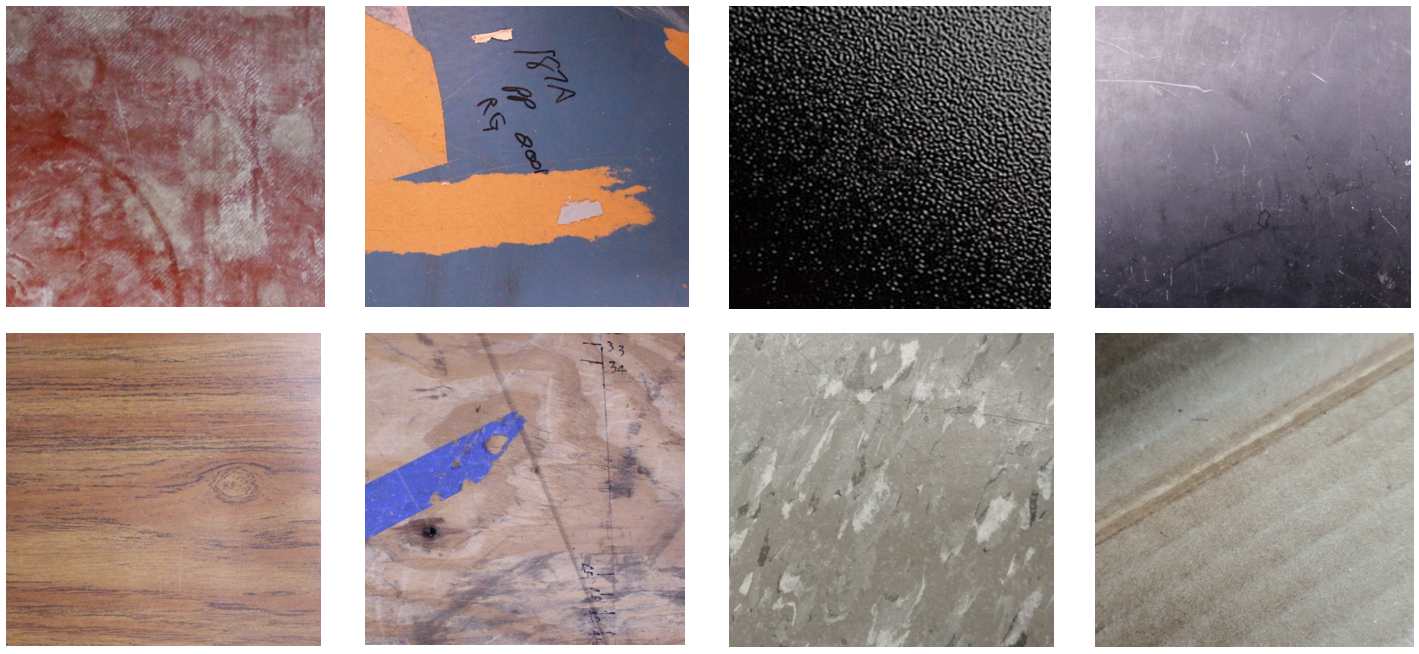}
\caption{Texture samples for the printing bed. More than 15 photographs of surfaces such as wood, metal, paper, stone, and others were superimposed on the virtual working area. Variations in lighting, cropping, scaling, and image orientation during animation allow the creation of unique backgrounds.}
\label{fig_5}
\end{figure}

The color of the plastic material and the surface characteristics of the printed part were adjusted empirically using a rich library of Blender shaders \cite{ref60}. When simulating surface irregularities, the Noise Texture \cite{ref61} and Voronoi Texture \cite{ref62} nodes were used to add Perlin and Worsley noises, respectively, while the "Bump" node was added to adjust the overall roughness. Photorealistic color, transparency, and reflection parameters were obtained by the combination of Principled \cite{ref63} (adds multiple layers to vary color, reflection, sheen, transmission, and other parameters), Glossy \cite{ref64} (adds reflection with microfacet distribution), Diffuse \cite{ref65} (adds Lambertian and Oren-Nayar diffuse reflections), and Transparent \cite{ref66} (adds transparency without refraction) Bidirectional Scattering Distribution Functions (BSDFs) (Figure 6).

\begin{figure}[!h]
\centering
\includegraphics[width=2.8in]{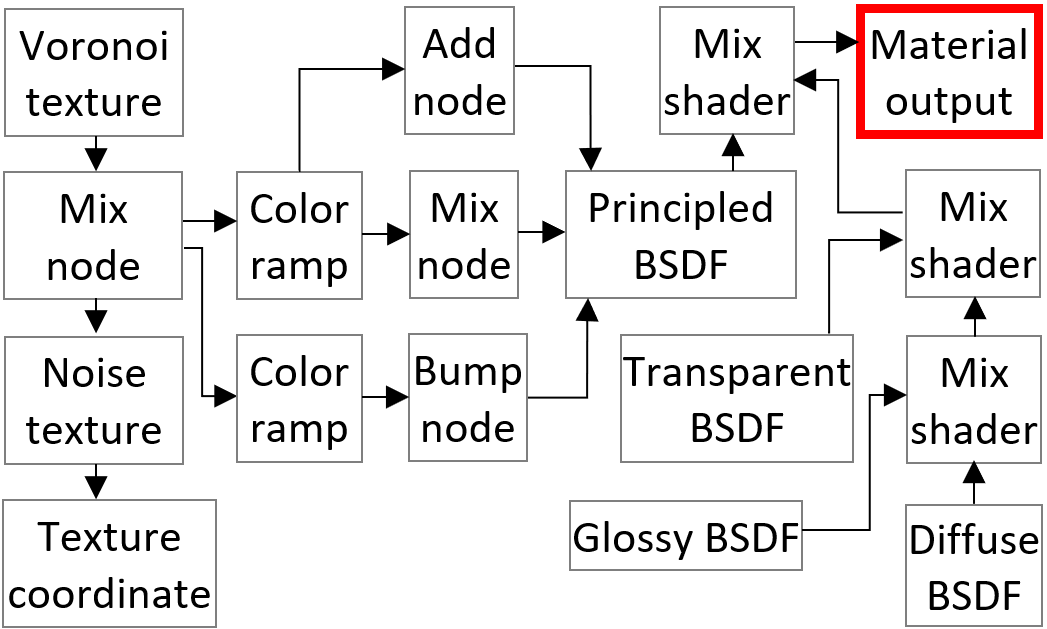}
\caption{Shading node network has been experimentally developed to achieve maximum realism of generated renders. The creation of all connections and node settings is fully automated in the code, which provides the flexibility to adjust the color, transparency, reflectivity, and other characteristics of the output material (red).}
\label{fig_6}
\end{figure}

The G-code parsing procedure heavily utilizes the functionality of the Blender application programming interface \cite{ref67}, which provides access to the properties of all shader nodes used in the scene. The entire animation process is scripted with randomized locations of the camera, light sources, and printing bed plane in timeline keyframes, where the graphics engine adds intermediate frames by interpolation. Most of the G-codes were used twice with different levels of part completion, material color, print surface texture, light source locations, and camera orientations, resulting in approximately100 unique synthetic images for each selected CAD design.

The built-in compositing interface \cite{ref68} was used to create pixel-perfect masks for each frame (Figure 7). During the slicing procedure, each extruder path acquires its own type, which can be visualized in pseudo colors in slicing environment (Figure 8). In this work, the outer and inner walls were combined into one structural element "shell". For visual segregation (masking) of individual scene elements (background, top layer, infill, shell, and support), different values of the material pass index parameter \cite{ref69} were set at the G-code parsing stage. This allows each selected element to be rendered as a region filled with a certain grayscale level.

\begin{figure}[!h]
\centering
\includegraphics[width=2.0in]{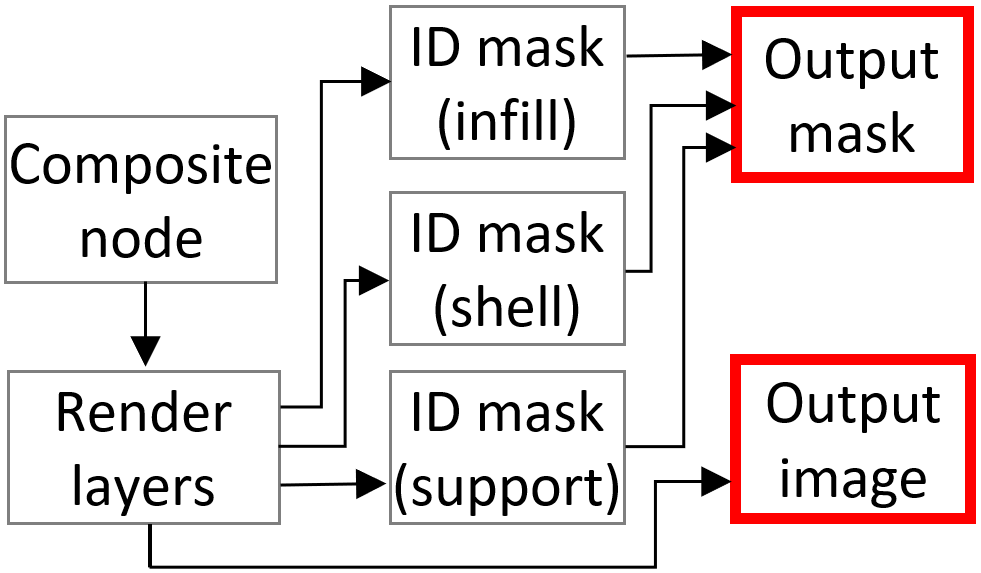}
\caption{Composite node network (internal structure segmentation example) assigns user-defined color labels to each pixel in the output image, depending on whether it belongs to a particular area (infill, shell, or support) of the rendered part. This creates a pixel-precise ground truth mask (red) for each output image frame (red) in the animation.}
\label{fig_7}
\end{figure}

The internal physics-based path tracer Cycles \cite{ref70} was used to render each frame of the animation. To reduce rendering time, the number of samples was set to 64, the total number of light path reflections was reduced to 8, and the Reflective and Refractive Caustics features were disabled. The Cycles performance depends on the system's computational power. An 8 GB GPU setup with 256x256 render tile size and an output image size of 1024x1024 pixels takes up to one minute to process a single frame, depending on the scale and geometric complexity of the scene within the camera viewport. Rendering an entire 50-frame animation this way can take up to one hour.

\begin{figure}[!h]
\subfloat[]{\includegraphics[height=0.8in]{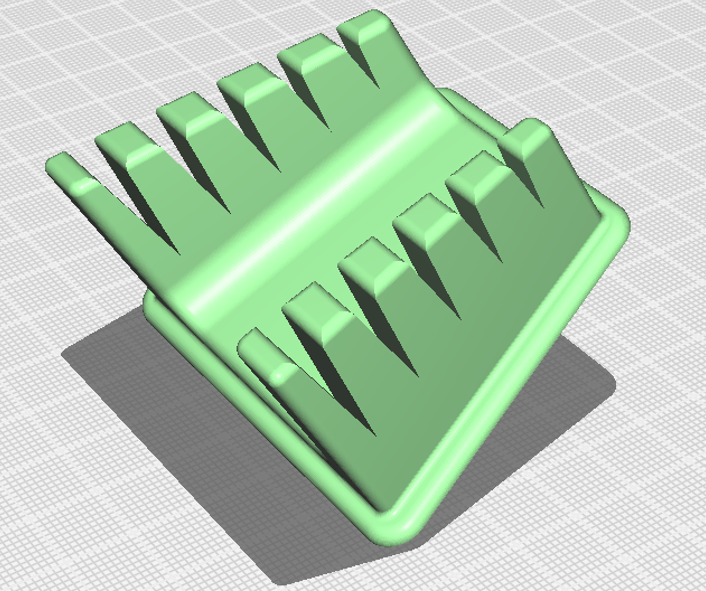}}
\hfil
\subfloat[]{\includegraphics[height=0.8in]{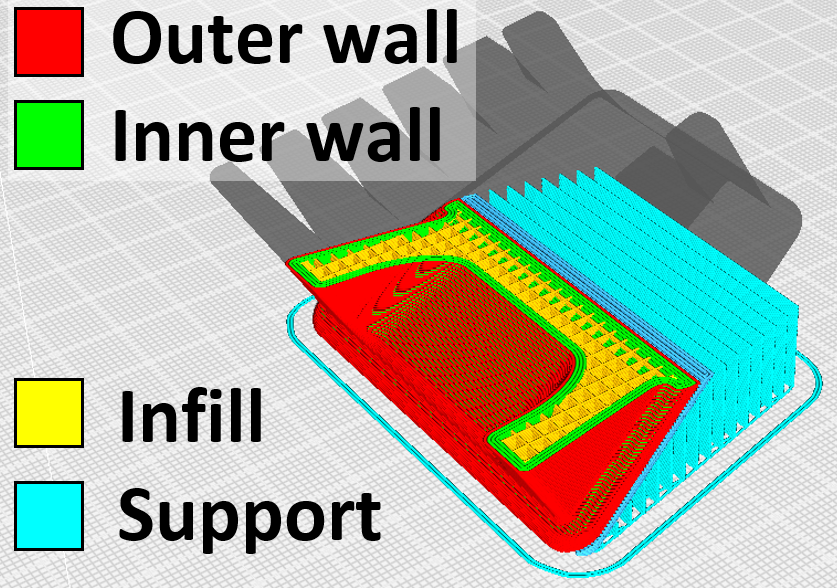}}
\hfil
\subfloat[]{\includegraphics[height=0.8in]{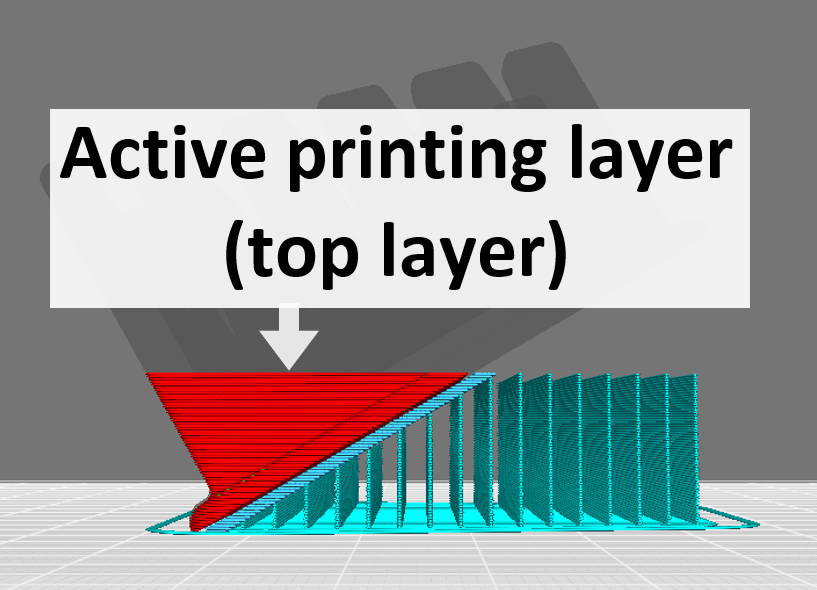}}
\caption{3-D model slicing procedure. (a) Whole part in STL format. (b) Internal structure of the sliced layers (red---outer shell, green---inner shell, yellow---infill, blue---support. (c) Side view illustrates the current printing layer (top layer at each manufacturing stage).}
\label{fig_8}
\end{figure}

\subsubsection{Synthetic image datasets}
For the further task of semantic segmentation, three separate datasets were created (Figure 9). A total of 5,763 1024x1024 pixels image-mask pairs were generated for the segmentation of the entire 3-D printed part, 3,570---for the top layer segmentation, and 1,140---for infill, shell, and support (internal layer structure) segmentation.

\begin{figure}[!h]
\subfloat[]{\includegraphics[height=2.4in]{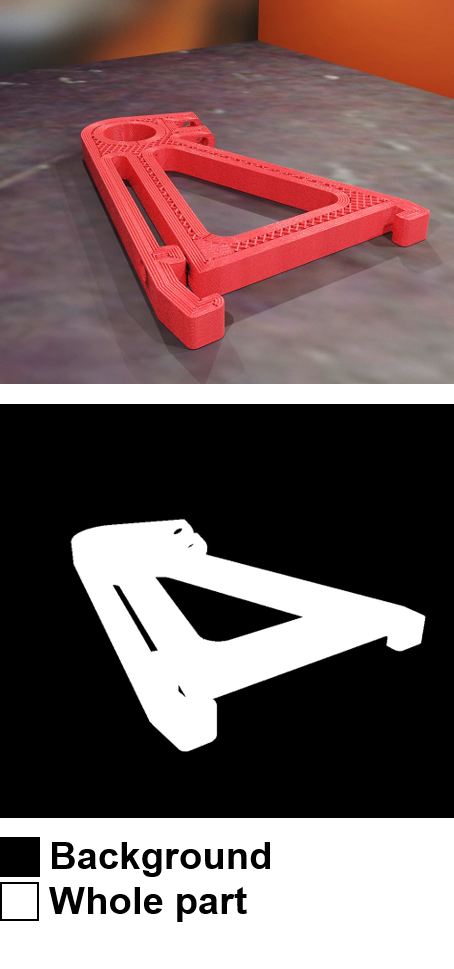}}
\hfil
\subfloat[]{\includegraphics[height=2.4in]{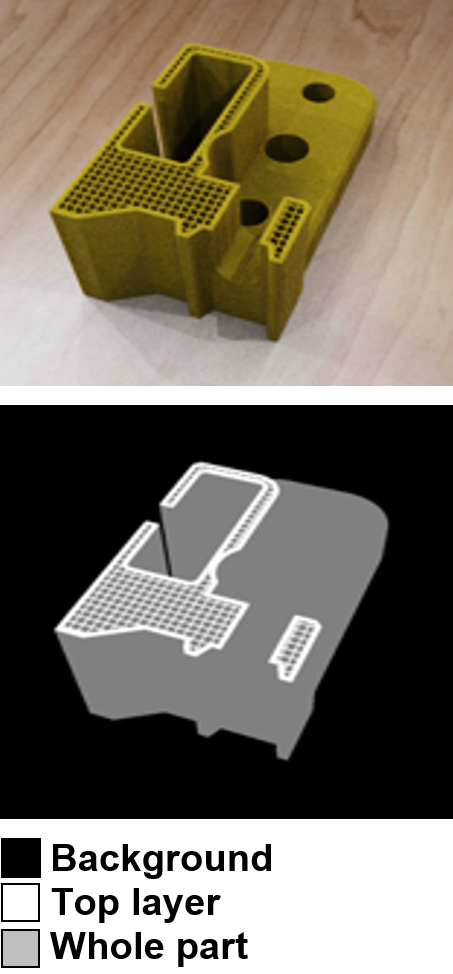}}
\hfil
\subfloat[]{\includegraphics[height=2.4in]{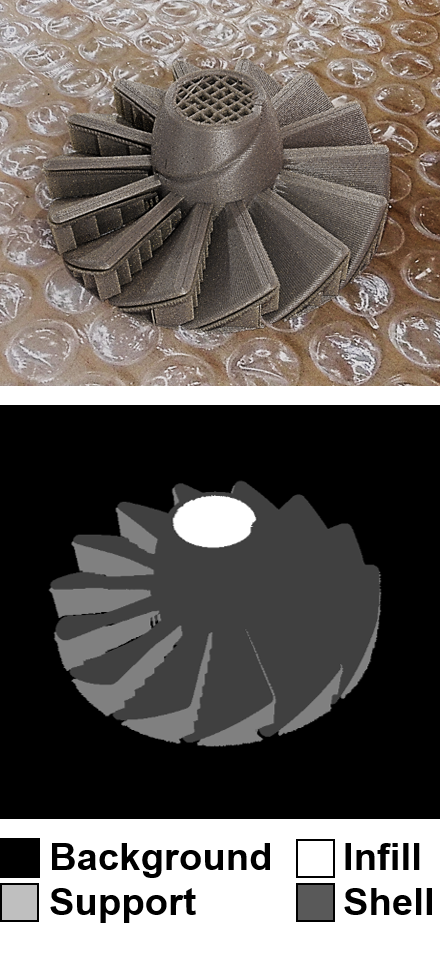}}
\caption{Image-mask pair samples for each AM synthetic dataset. (a) Whole part segmentation. (b) Top layer segmentation. (c) Internal layer segmentation.}
\label{fig_9}
\end{figure}

\subsection{Semantic image segmentation}
Minaee et al. \cite{ref32} and Ulku \& Akagündüz \cite{ref40} presented a comprehensive overview of the modern research state in the field of semantic segmentation.  As can be seen from the works \cite{ref71,ref72,ref73}, the U-Net family of neural network architectures has demonstrated high segmentation efficiency with small amounts of training data. The DeepLab architecture, in turn, is one of the basic architectures for subsequent domain adaptation \cite{ref74,ref75,ref76}.

This work employs the U-Net architecture \cite{ref77} and its multi-class adaptation \cite{ref78} due to its efficiency and simplicity.

\subsection{Image-to-image translation}
To potentially improve the efficiency of semantic segmentation, the application of the unpaired image-to-image translation method based on the CycleGAN network \cite{ref39} was considered. The given method learns the mapping between the source domain (real images) and the target domain (synthetic images) by minimizing the cycle consistency loss {\it{L$_{\text{C}}$}} (Figure 10) in the absence of paired data samples. 
 
\begin{figure}[!h]
\centering
\includegraphics[width=2.8in]{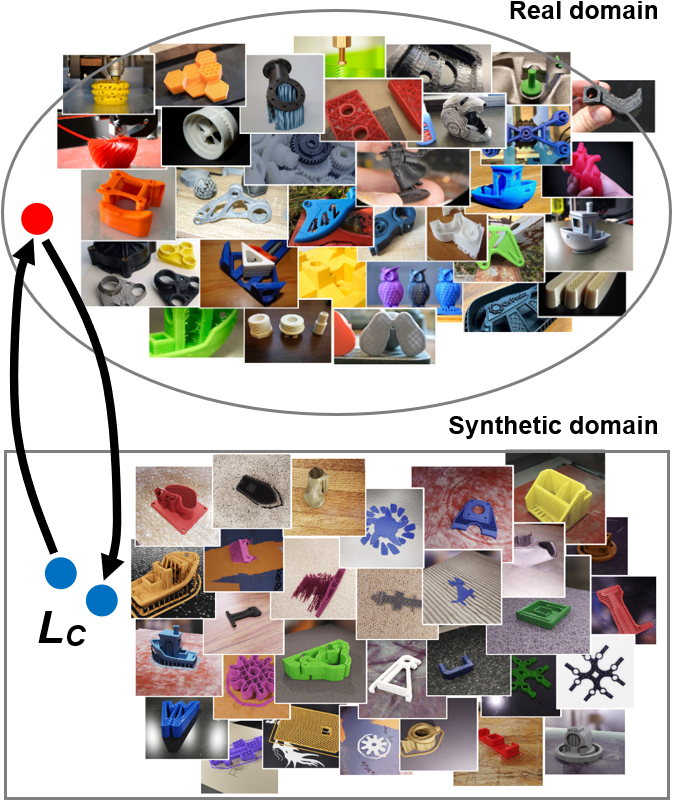}
\centering
\caption{Unpaired image-to-image translation using the cycle-consistent adversarial network. Handpicked images of real and virtual printed parts were loaded into CycleGAN, which learns to map real domain images to their synthetic counterparts and vice versa, minimizing the cycle consistency loss {\it{L$_{\text{C}}$}}.}
\label{fig_10}
\end{figure}

For this task were manually selected 589 synthetic renders and 794 real images of 3-D printed parts. The learning result is two generators that convert the original images of the real domain into their synthetic counterparts, and vice versa (Figure 11).

\begin{figure}[!h]
\centering
\includegraphics[width=3.2in]{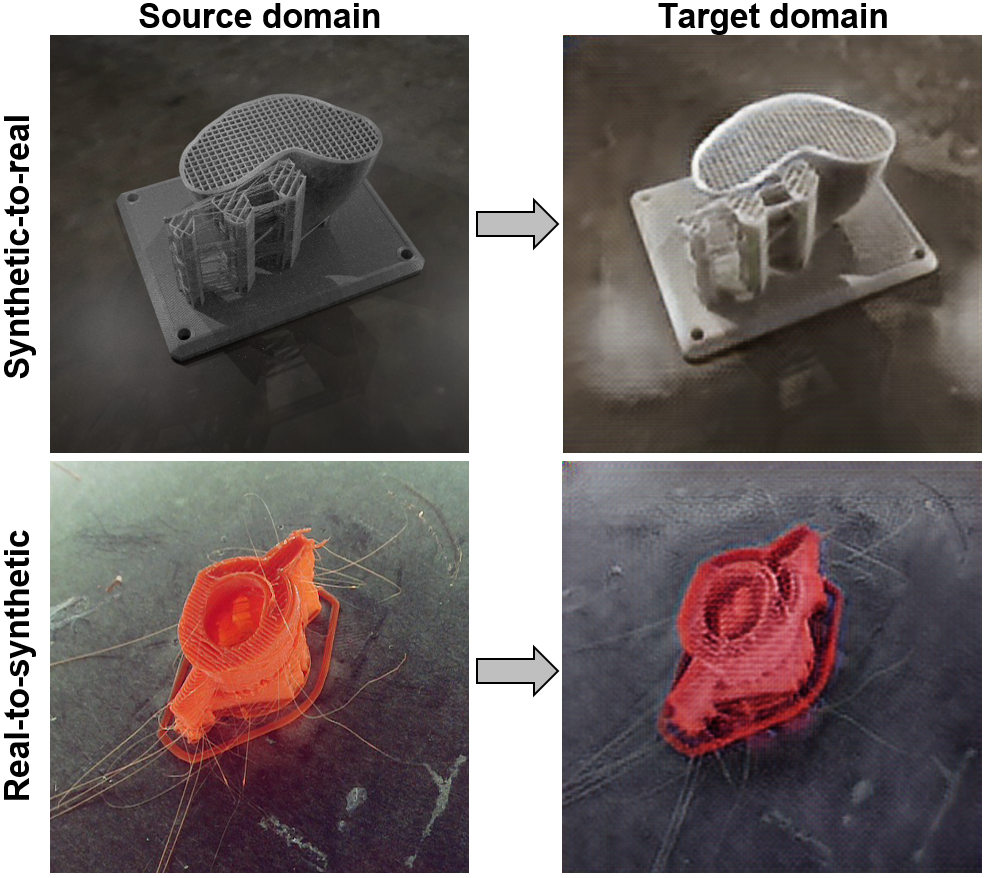}
\caption{Image-to-image style translation example. Translating a real image into its synthetic version reduces the contrast and saturation of the reflections and incidental filament strings.}
\label{fig_11}
\end{figure}

As can be seen from Figure 11, translating a synthetic render into a real image makes colors more natural, while translating a real image to a synthetic one also reduces the contrast and saturation of the reflections on the printing bed and incidental filament strings. This characteristic can improve segmentation in mediocre images.

\section{Results}
The results of semantic segmentation are presented on the example of several real images in Figure 12. The training of the neural network was carried out on synthetic renders without using the style translation technique.

Quantitative results are shown in Table 1. Test datasets include synthetic renders of STL models both included in the training dataset and not included in it. 3-D Models included in the training dataset have their color, angle, and environment parameters changed to avoid matching the data the model was trained on.

\begin{figure*}[!b]
\centering
\includegraphics[width=5.8in]{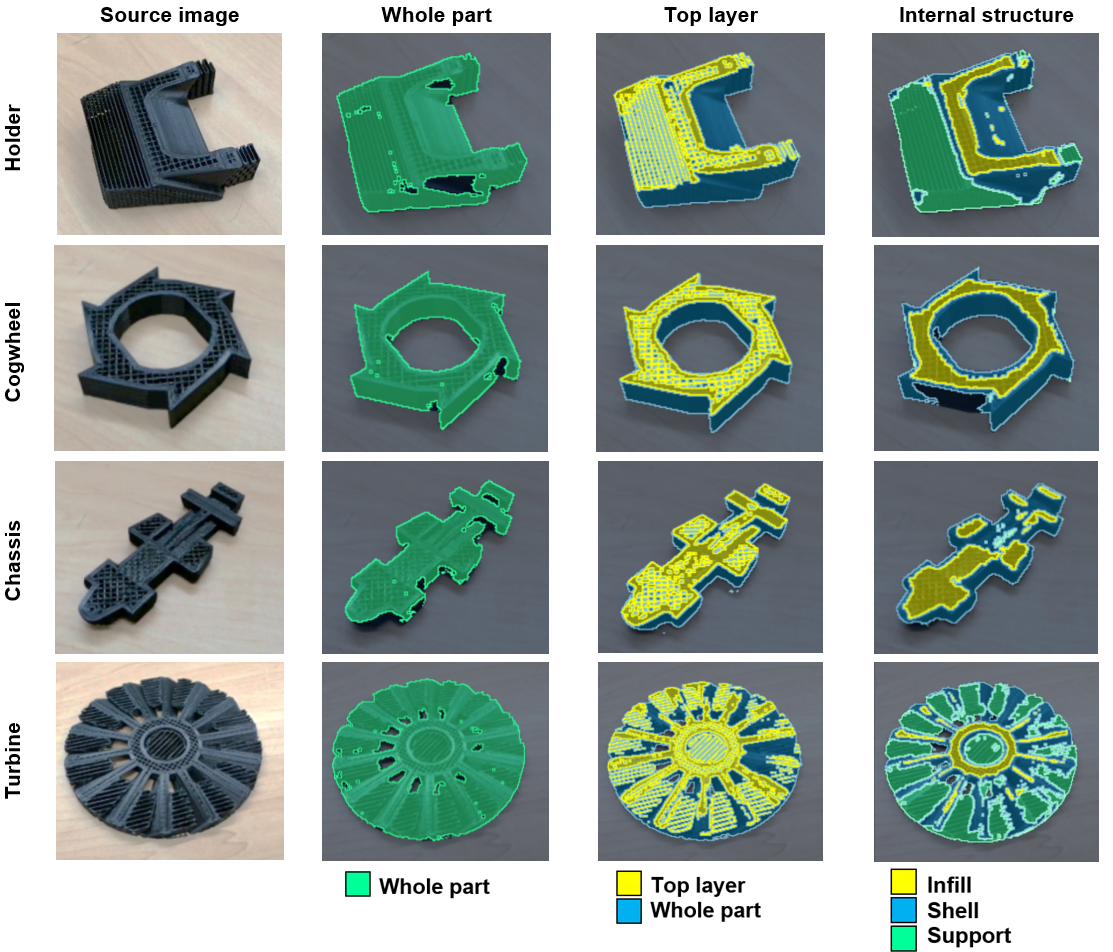}
\caption{Results of semantic segmentation on the example of several real images. The neural network was trained on similar synthetic 3-D models. Color, printing surface texture, and slicing parameters, however, differ from those used in the training dataset.}
\label{fig_12}
\end{figure*}

\begin{table*}[!b]
\caption{Segmentation results for synthetic test datasets (mIoU scores, \%)\label{tab:table1}}
\centering
\begin{tabular}{|c|c|c|c|c|c|c|}
\hline
\textbf{No. of images} & \textbf{Test dataset} & \textbf{Background} & \textbf{Top layer} & \textbf{Shell} & \textbf{Support} & \textbf{Infill}\\
\hline
89 &  Whole part segmentation (real images) & 78.16 & --- & --- & --- & --- \\
101 & Whole part segmentation (synthetic renders images) & 94.90 & --- & --- & --- & --- \\
68 & Top layer segmentation (synthetic renders) & 99.74 & 73.33 & --- & --- & --- \\
57 & Internal structure segmentation (synthetic renders) & 94.52 & --- & 55.31 & 69.45 & 78.93 \\
 \hline
\end{tabular}
\end{table*}

The intersection over union (IoU) quantifies the degree of overlap (from 0 to 100\%) between the ground truth mask and the segmented pixel area of its predicted version, where a larger value indicates a more accurate segmentation and the mIoU is the mean IoU value across the correspondent classes in the dataset. The calculation of mIoU scores for real images was carried out only for the segmentation of the entire part, since the obtaining manually-labeled ground truth masks for the top layer and the internal structure of the part is a non-trivial task, considering the geometric complexity of the filling elements.

As can be seen from Table 1, the segmentation accuracy on real images (78.16\%) is inferior to synthetic data (94.90\%), which indicates the need for additional research on domain adaptation. Detecting the top layer is a more complex task for the neural network compared to segmenting the entire part, which is clearly noticeable in the results within the same dataset (mIoU 73.33\% for the top layer versus 99.74\% for the background). Shell segmentation has the lowest score (mIoU 55.31\%). This, apparently, is due to the variety of geometric shapes and the lack of a characteristic texture that the infill and support areas have.

To analyze the influence of style transfer (ST) on semantic segmentation, a separate CNN training of three datasets of one part was carried out (Figure 13). Synthetic and real datasets consist of 49 and 36 image-mask pairs, respectively.

\begin{figure}[!h]
\centering
\subfloat[]{\includegraphics[height=2.0in]{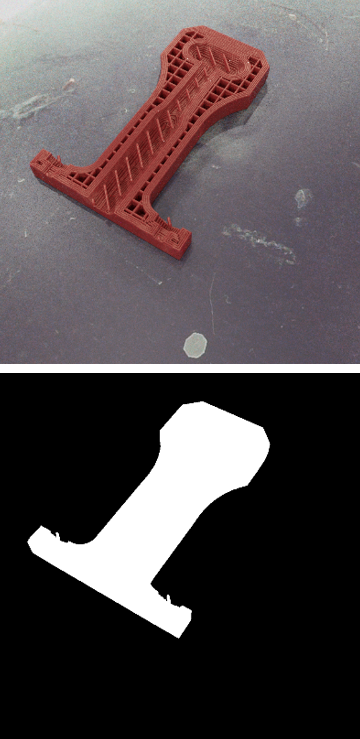}}
\hfil
\subfloat[]{\includegraphics[height=2.0in]{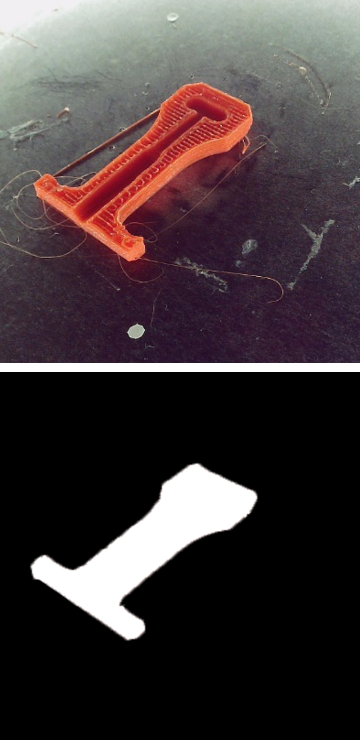}}
\hfil
\subfloat[]{\includegraphics[height=2.0in]{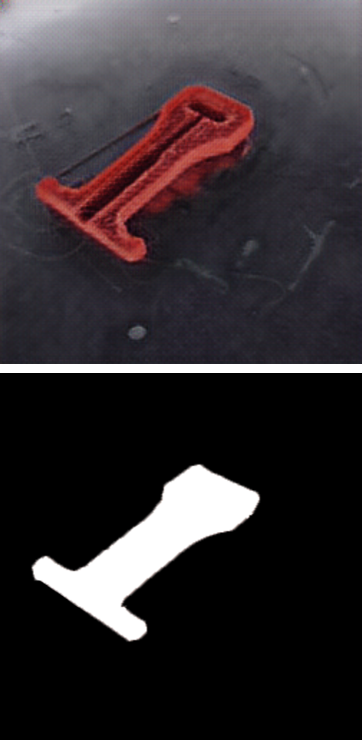}}
\caption{Datasets for the style transfer influence analysis. (a) Synthetic data. (b) Real data. (c) Real data after style transfer. The upper row shows sample images and the lower row illustrates the corresponding ground truth masks.}
\label{fig_13}
\end{figure}

To compare the domains were used t-distributed stochastic neighbor embedding (t-SNE) \cite{ref79,ref80} projections of the normalized bottleneck layers of trained U-Net models (Figure 14). The nonlinear dimensionality reduction technique was applied to 512-dimensional normalized vectors in the narrowest parts of the trained models to visualize the affinity of the domains in latent feature space. As can be seen from Figure 14a, the feature space of the real domain (orange) is getting closer to synthetic data (blue) after the image-to-image style translation (black).

\begin{figure}[!h]
\subfloat[]{\includegraphics[height=1.3in]{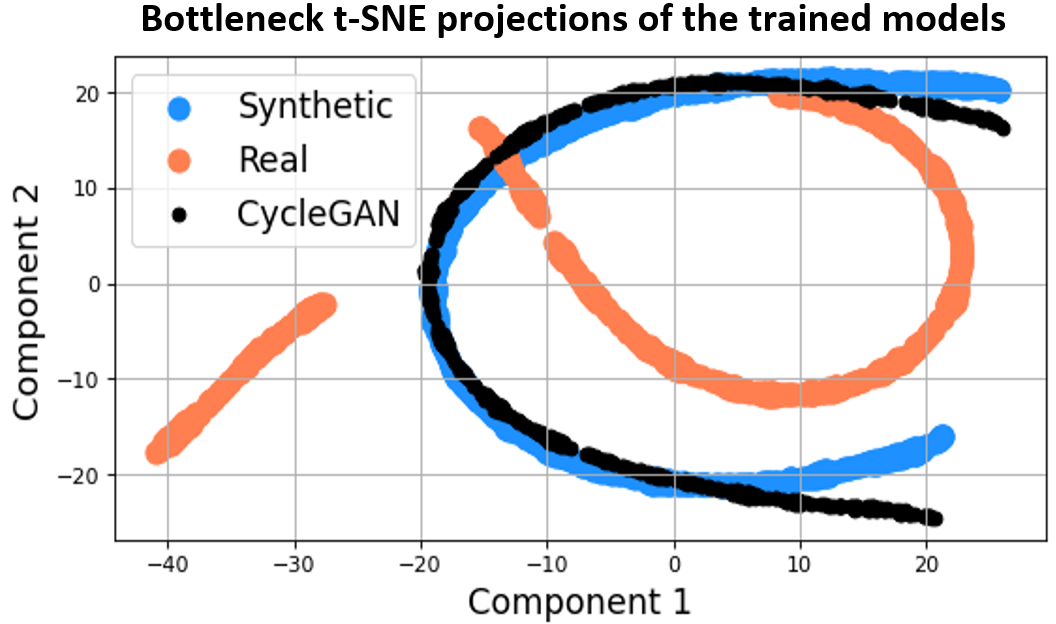}}
\hfil
\subfloat[]{\includegraphics[height=1.3in]{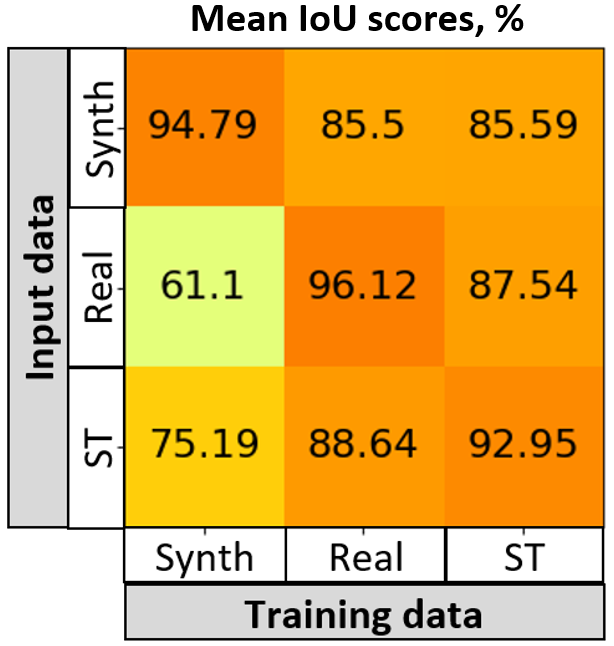}}
\caption{Domain comparison via t-SNE projections (a) and segmentation
performance before and after style translation (b).}
\label{fig_14}
\end{figure}

In addition to t-SNE projections, the segmentation performance of the source real image data after ST was also analyzed (Figure 14b). The heatmap columns represent the data on which the neural network model was trained, and the rows stand for the input data on which segmentation was applied.  The highest mIoU, as expected, was observed in those data sets on which the model was trained. When converting the real input data into ST using the image-to-image translation, however, the segmentation score increased from 61.10\% to 75.19\% for the model trained solely on synthetic data. This parameter is the most valuable, since in real conditions, training a convolutional network on real data may not be possible due to the lack of ground truth masks. This indicates that the ST method as a precursor to domain adaptation can significantly improve real 3-D printing image segmentation in situations where a model trained on synthetic data will be the only tool available. The sample results of image segmentation before and after style translation are shown in Figure 15.

\begin{figure*}[!h]
\centering
\includegraphics[width=5.6in]{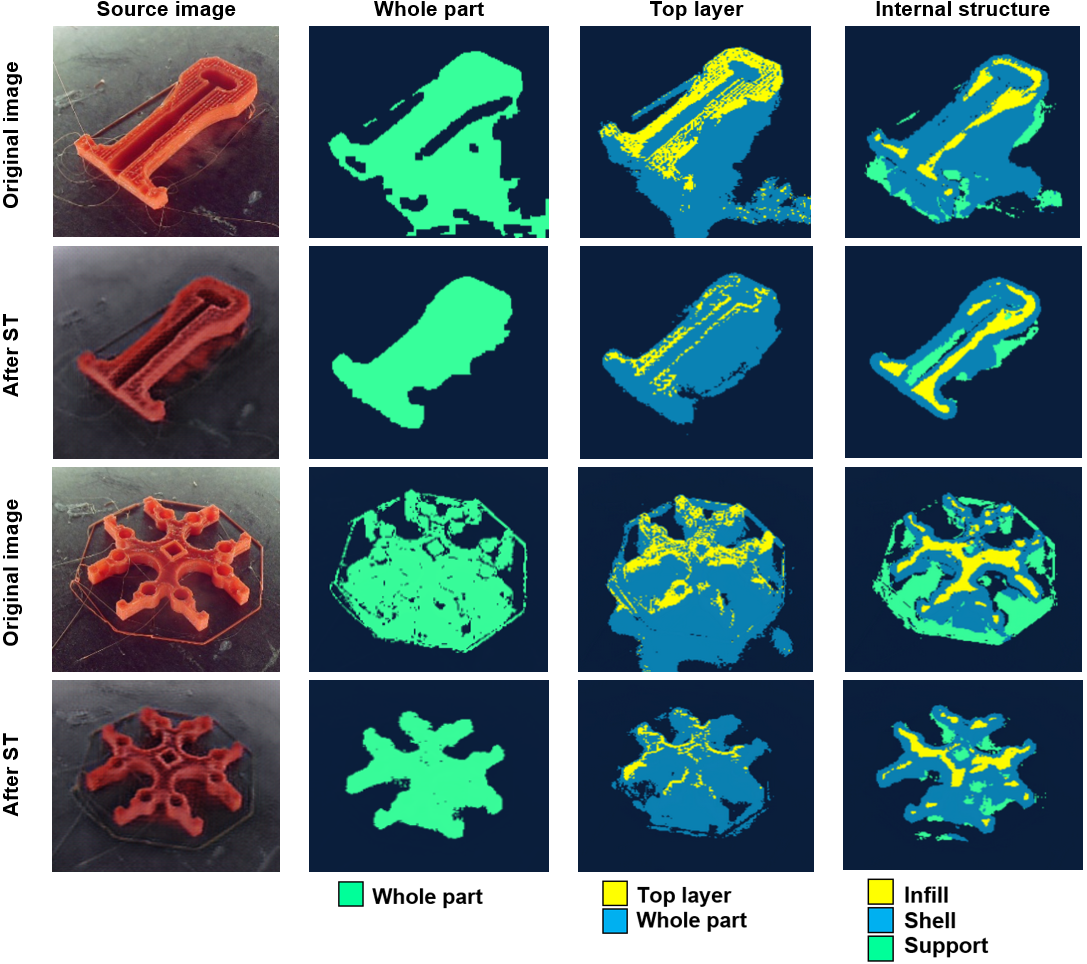}
\caption{Results of image segmentation before and after style translation. Real-to-synthetic style transfer reduces the saturation of the incidental filament strings and reflections on the printing platform, which, in turn, affects the results of semantic segmentation.}
\label{fig_15}
\end{figure*}

As can be seen from Figure 15, real-to-synthetic style transferring reduces the saturation of the incidental filament strings and reflections on the printing platform, which, in turn, affects the results of semantic segmentation. Image-to-image translation, therefore, could be a powerful tool in further improving segmentation performance through domain adaptation techniques.

This work continues the previous authors' research on the use of physical rendering and demonstrates the significant potential of using synthetic data and machine learning in the field of additive manufacturing. Due to the relative simplicity of virtual printing and training data generation, segmentation of the contours of a manufactured part can be performed at every stage of its completion using a single camera in an arbitrary position. This reduces the requirements for camera calibration and eliminates the need to use visual markers to tightly bind the image frame to the coordinate system of the 3-D printing space. It also offers the flexibility to be used on any type of 3-D printing system with the addition of an after-market camera.

The limitations of the developed method are the need to create synthetic images and increase the training dataset for each new manufacturing part, as well as the implementation of transfer learning to improve the segmentation accuracy. Additional research is also required in the field of domain adaptation application based on existing state-of-the-art techniques \cite{ref81,ref82,ref83}.

Together with edge-based markerless tracking \cite{ref84,ref85}, the developed technique can become an integral part of a 3-D printing control and monitoring system such as OctoPrint \cite{ref86}. In the future, this will make it possible to implement an in-line comprehensive system for recognizing the type of a part being produced and determining its location and orientation in the workspace, as well as for tracking its manufacturing deviations.

\section{Conclusions}
The semantic segmentation framework for additive manufacturing can enhance the visual analysis of manufacturing processes and allow the detection of individual manufacturing errors, while significantly reducing the requirements for positioning accuracy and camera calibration.

The results of this work will allow localizing of 3-D printed parts in captured image frames and applying image processing techniques to its structural elements for following tracking of manufacturing deviations. The use of image style transfer is of significant value for further research in the field of adapting the domain of synthetic renders to real images of 3-D printed products.

The methodology demonstrated achieved the following mIoU scores for synthetic test datasets: entire printed part 94.90\%, top layer 73.33\%, infill 78.93\%, shell 55.31\%, support 69.45\%. The results illustrate the effectiveness of the developed method, but also indicate the need for additional experiments to eliminate the synthetic-to-real domain gap.

\section*{Acknowledgments}
The authors would like to thank Heinz Löpmeier for permission to use his code as a basis template for G-code parsing, as well as Doug Everett and Kenneth Jiang for access to the Spaghetti Detective’s user performance database. This work was supported by the Thompson Endowments and the Natural Sciences and Engineering Research Council of Canada.

\section*{Biography Section}

\vspace{-30pt}

\begin{IEEEbiographynophoto}{Aliaksei Petsiuk}
is a PhD candidate at University of Western Ontario (Canada) specializing in 3-D printing and additive manufacturing. He is a member of Tau Beta Pi, the engineering honor society, the managing editor of \emph{HardwareX} journal, and a reviewer of \emph{Additive Manufacturing} journal. He is highly passionate about combining computer vision and augmented reality technologies with intelligent sensor systems in solving manufacturing problems.
\end{IEEEbiographynophoto}

\vspace{-30pt}

\begin{IEEEbiographynophoto}{Harnoor Singh}
is an MEng graduate student of University of Western Ontario focused on image processing and machine learning.
\end{IEEEbiographynophoto}

\vspace{-30pt}

\begin{IEEEbiographynophoto}{Himanshu Dadhwal}
is an MEng graduate student of University of Western Ontario specializing in computer engineering.
\end{IEEEbiographynophoto}

\vspace{-30pt}

\begin{IEEEbiographynophoto}{Joshua M. Pearce}
is the John M. Thompson Chair in Information Technology and Innovation at the Thompson Centre for Engineering Leadership \& Innovation (University of Western Ontario, Canada). His research concentrates on the use of open source appropriate technology (OSAT) to find collaborative solutions to problems in sustainability and to reduce poverty. His research spans areas of engineering of solar photovoltaic technology, open hardware, and distributed recycling and additive manufacturing (DRAM) using RepRap 3-D printing, but also includes policy and economics. He is the editor-in-chief of \emph{HardwareX}, the first journal dedicated to open source scientific hardware and the author of the \emph{Open-Source Lab: How to Build Your Own Hardware and Reduce Research Costs}. He is also the author of \emph{Create, Share, and Save Money Using Open-Source Projects}, and \emph{To Catch the Sun}, an open source book of inspiring stories of communities coming together to harness their own solar energy.
\end{IEEEbiographynophoto}

\vspace{11pt}

\vfill


\begin{thebibliography}{1}
\bibliographystyle{IEEEtran}

\bibitem{ref1}	R. Geyer, J.R. Jambeck, and K.L. Law, “Production, Use, and Fate of all Plastics Ever Made,” {\it{Sci. Adv.,}} 2017, 3(7), e1700782, doi: 10.1126/sciadv.1700782.

\bibitem{ref2}	J.R. Jambeck, R. Geyer, C. Wilcox, T.R. Siegler, M. Perryman, A. Andrady, R. Narayan, and K.L. Law, “Plastic Waste Inputs From Land Into the Ocean,” {\it{Science,}} 2015, 347(6223), pp. 768–771, doi: 10.1126/science.1260352.

\bibitem{ref3}	A.O. Laplume, B. Petersen, and J.M. Pearce, "Global value chains from a 3D printing perspective," {\it{J Int Bus Stud,}} 2016, 47(5), pp. 595-609, doi: 10.1057/jibs.2015.47.

\bibitem{ref4}	E.E. Petersen and J.M. Pearce, "Emergence of home manufacturing in the developed world: Return on investment for open-source 3-D printers," {\it{Technologies,}} 2017, 5(1), 7.

\bibitem{ref5}	J.M. Pearce and J.Y. Qian, "Economic Impact of DIY Home Manufacturing of Consumer Products with Low-cost 3D Printing from Free and Open Source Designs," {\it{European Journal of Social Impact and Circular Economy,}} 2022, 3(2), 1-24, doi: 10.13135/2704-9906/6508.

\bibitem{ref6}	E. Hunt, C. Zhang, N. Anzalone, and J.M. Pearce, "Polymer recycling codes for distributed manufacturing with 3-D printers," {\it{Resources, Conservation and Recycling,}} 2015, 97, 24-30.

\bibitem{ref7}	N. Shahrubudin, T.C. Lee, and R. Ramlan, "An overview on 3D printing technology: Technological, materials, and applications," {\it{Procedia Manufacturing,}} 2019, 35, 1286-1296.

\bibitem{ref8}	"Global 3D Printing Filament Market By Material, By Type, By End Use, By Region, Competition, Forecast \& Opportunities, 2024," May, 2019. Available: https://www.reportbuyer.com/product/5778909/global-3d-printing-filament-market-by-material-by-typeby-end-use-by-region-competition-forecast-and-opportunities-2024.html. [Accessed Oct. 10, 2022].

\bibitem{ref9}	B.T. Wittbrodt, A.G. Glover, J. Laureto, G.C. Anzalone, D. Oppliger, J.L. Irwin, and J.M. Pearce, "Life-cycle economic analysis of distributed manufacturing with open-source 3-D printers," {\it{Mechatronics,}} 2013, 23(6), 713-726.

\bibitem{ref10}	S. Sharp, CEO 3DQue Systems. Personal communication. June 4, 2022.

\bibitem{ref11}	H.D. Kang, "Analysis of furniture design cases using 3D printing technique," {\it{The Journal of the Korea Contents Association,}} 2015, 15(2), 177-186.

\bibitem{ref12}	J.K. Bow, N. Gallup, S.A. Sadat, and J.M. Pearce, "Open source surgical fracture table for digitally distributed manufacturing," {\it{PloS one,}} 2022, 17(7), e0270328.

\bibitem{ref13}	J.I. Novak and J. O’Neill, "A design for additive manufacturing case study: fingerprint stool on a BigRep ONE," {\it{Rapid Prototyping Journal,}} 2019, 25(6), pp. 1069-1079.

\bibitem{ref14}	A. Petsiuk, B. Lavu, R. Dick, and J.M. Pearce, "Waste Plastic Direct Extrusion Hangprinter," {\it{Inventions,}} 2022, 7(3), p.70.

\bibitem{ref15}	A.L. Woern, D.J. Byard, R.B. Oakley, M.J. Fiedler, S.L. Snabes, and J.M. Pearce, "Fused particle fabrication 3-D printing: Recycled materials’ optimization and mechanical properties," {\it{Materials,}} 2018, 11(8), p.1413.

\bibitem{ref16}	A. Oleff, B. Küster, M. Stonis, L. Overmeyer, "Process monitoring for material extrusion additive manufacturing: a state-of-the-art review," {\it{Prog Addit Manuf,}} 2021, 6, pp. 705-730, doi: 10.1007/s40964-021-00192-4.

\bibitem{ref17}	A. Ceruti, A. Liverani, T. Bombardi, "Augmented vision and interactive monitoring in 3D printing process," {\it{Int J Inter Des Manuf,}} 2017, 11, pp. 385-395, doi: 10.1007/s12008-016-0347-y.

\bibitem{ref18}	S. Nuchitprasitchai, M.C. Roggemann, J.M. Pearce, "Factors effecting real-time optical monitoring of fused filament 3D printing," {\it{Prog Addit Manuf J,}} 2017, 2(3), pp. 133-149, doi: 10.1007/s40964-017-0027-x.

\bibitem{ref19}	A. Johnson, H. Zarezadeh, X. Han, R. Bibb, R. Harris, "Establishing in-process inspection requirements for material extrusion additive manufacturing," in {\it{Proceedings of the Fraunhofer Direct Digital Manufacturing Conference. Berlin: Fraunhofer-Gesellschaft, 2016.}}

\bibitem{ref20}	S. Hurd, C. Camp, J. White, "Quality assurance in additive manufacturing through mobile computing," {\it{Int Conf Mob Comput Appl Serv,}} 2015, pp. 203-220.

\bibitem{ref21}	H. Jeong, M. Kim, B. Park, S. Lee, "Vision-Based Real-Time Layer Error Quantification for Additive Manufacturing," in {\it{Proc ASME 2017 12th Int Manuf Sci Eng Conf, Los Angeles, California, USA, 2017.}}

\bibitem{ref22}	F. Wasserfall, D. Ahlers and N. Hendrich, "Optical In-Situ Verification of 3D-Printed Electronic Circuits," {\it{IEEE 15th Int Conf Autom Sci and Eng (CASE),}} 2019, pp. 1302-1307, doi: 10.1109/COASE.2019.8842835.

\bibitem{ref23}	J. Straub, "3D printing cybersecurity: Detecting and preventing attacks that seek to weaken a printed object by changing fill level," {\it{Proc SPIE Dimens Opt Metrol Insp Pract Appl VI,}} 2017, doi: 10.1117/12.2264575.

\bibitem{ref24}	M.D. Kutzer, L.D. DeVries, C.D. Blas, "Part monitoring and quality assessment of conformal additive manufacturing using image reconstruction," in {\it{Proc ASME 2018 Int Des Eng Tech Conf Comput Inf Eng Conf 5B, Quebec, Canada, 2018,}} doi: 10.1115/DETC2018-85370.

\bibitem{ref25}	Z. Chen and R. Horowitz, "Vision-assisted Arm Motion Planning for Freeform 3D Printing," {\it{2019 American Control Conference (ACC),}} pp. 4204-4209, doi: 10.23919/ACC.2019.8814699.

\bibitem{ref26}	H. Shen, W. Du, W. Sun, Y. Xu, J. Fu, "Visual detection of surface defects based on self-feature comparison in robot 3-D printing," {\it{Appl Sci,}} 2020, 10(1), 235, doi: 10.3390/app10010235.

\bibitem{ref27}	A. Malik, H. Lhachemi, J. Ploennigs, A. Ba, R. Shorten, "An application of 3D model reconstruction and augmented reality for real-time monitoring of additive manufacturing," {\it{Procedia CIRP,}} 2019, 81, pp. 346-351.

\bibitem{ref28}	A. Petsiuk, J.M. Pearce, "Open source computer vision-based layer-wise 3D printing analysis," {\it{Addit Manuf,}} 2020, 36, 101473, doi: 10.1016/j.addma.2020.101473.

\bibitem{ref29}	A. Petsiuk, J.M. Pearce, "Towards smart monitored AM: Open source in-situ layer-wise 3D printing image anomaly detection using histograms of oriented gradients and a physics-based rendering engine," {\it{Addit Manuf,}} 2022, 52, 102690, doi: 10.1016/j.addma.2022.102690.

\bibitem{ref30}	Spaghetti Detective. Available:  https://www.obico.io/the-spaghetti-detective.html. [Accessed Oct. 10, 2022].

\bibitem{ref31}	The Spaghetti Detective Plugin. Available: https://github.com/TheSpaghettiDetective/OctoPrint-TheSpaghettiDetective. [Accessed Oct. 10, 2022].

\bibitem{ref32}	S. Minaee, Y. Boykov, F. Porikli, A. Plaza, N. Kehtarnavaz and D. Terzopoulos, "Image Segmentation Using Deep Learning: A Survey," in {\it{IEEE Transactions on Pattern Analysis and Machine Intelligence,}} 2022, 44(7), pp. 3523-3542, doi: 10.1109/TPAMI.2021.3059968.

\bibitem{ref33}	Blender: the free and open source 3D creation suite. Available: https://www.blender.org. [Accessed Oct. 10, 2022].

\bibitem{ref34}	G. Csurka, R. Volpi, B. Chidlovskii, "Unsupervised Domain Adaptation for Semantic Image Segmentation: a Comprehensive Survey," 2021, {\it{arXiv:2112.03241.}}

\bibitem{ref35}	A. Farahani, S. Voghoei, K. Rasheed, H.R. Arabnia, "A Brief Review of Domain Adaptation," in {\it{Advances in Data Science and Information Engineering. Transactions on Computational Science and Computational Intelligence. Springer, Cham, 2021,}} doi: 10.1007/978-3-030-71704-9\_65.

\bibitem{ref36}	B. Imbusch, M. Schwarz, S. Behnke, "Synthetic-to-Real Domain Adaptation using Contrastive Unpaired Translation," 2022, {\it{arXiv:2203.09454.}}

\bibitem{ref37}	P. Li, X. Liang, D. Jia, E.P. Xing, "Semantic-aware Grad-GAN for Virtual-to-Real Urban Scene Adaption," 2018, {\it{arXiv:1801.01726.}}

\bibitem{ref38}	S. Lee, E. Park, H. Yi, S.H. Lee, "StRDAN: Synthetic-to-Real Domain Adaptation Network for Vehicle Re-Identification," in {\it{IEEE Conference on Computer Vision and Pattern Recognition (CVPR), 2020.}}

\bibitem{ref39}	J. -Y. Zhu, T. Park, P. Isola and A. A. Efros, "Unpaired Image-to-Image Translation Using Cycle-Consistent Adversarial Networks," in {\it{2017 IEEE International Conference on Computer Vision (ICCV),}} pp. 2242-2251, doi: 10.1109/ICCV.2017.244.

\bibitem{ref40}	I. Ulku and E. Akagündüz, "A Survey on Deep Learning-based Architectures for Semantic Segmentation on 2D Images," {\it{Applied Artificial Intelligence,}} 2022, 36(1), doi: 10.1080/08839514.2022.2032924.

\bibitem{ref41}	M. Cordts, M. Omran, S. Ramos, T. Rehfeld, M. Enzweiler, R. Benenson, U. Franke, S. Roth, B. Schiele, "The Cityscapes Dataset for Semantic Urban Scene Understanding," in {\it{IEEE Conference on Computer Vision and Pattern Recognition (CVPR), 2016.}}

\bibitem{ref42}	S.R. Richter, V. Vineet, S. Roth, V. Koltun, "Playing for Data: Ground Truth from Computer Games," 2016, {\it{arXiv:1608.02192.}}

\bibitem{ref43}	G. Ros, L. Sellart, J. Materzynska, D. Vazquez and A. M. Lopez, "The SYNTHIA Dataset: A Large Collection of Synthetic Images for Semantic Segmentation of Urban Scenes," in {\it{2016 IEEE Conference on Computer Vision and Pattern Recognition (CVPR),}} pp. 3234-3243, doi: 10.1109/CVPR.2016.352.

\bibitem{ref44}	S.I. Nikolenko. Synthetic Data for Deep Learning. SOIA, vol. 174. {\it{Springer, Cham,}} 2021, doi: 10.1007/978-3-030-75178-4.

\bibitem{ref45}	C.M. de Melo, A. Torralba, L. Guibas, J. DiCarlo, R. Chellappa, J. Hodgins, "Next-generation deep learning based on simulators and synthetic data," {\it{Trends Cogn Sci,}} 2022, 26(2), pp. 174-187, doi: 10.1016/j.tics.2021.11.008.

\bibitem{ref46}	D. Ward, P. Moghadam, N. Hudson, "Deep Leaf Segmentation Using Synthetic Data," 2018, {\it{arXiv:1807.10931.}}

\bibitem{ref47}	A. Boikov, V. Payor, R. Savelev, A. Kolesnikov, "Synthetic Data Generation for Steel Defect Detection and Classification Using Deep Learning," {\it{Symmetry,}} 2021, 13(7):1176, doi: 10.3390/sym13071176.

\bibitem{ref48}	M. Valizadeh, S.J. Wolff, "Convolutional Neural Network applications in additive manufacturing: A review," {\it{Adv in Ind and Manuf Eng,}} 2022, 4, 100072, doi: 10.1016/j.aime.2022.100072.

\bibitem{ref49}	Y. Banadaki, N. Razaviarab, H. Fekrmandi, S. Sharifi, "Toward Enabling a Reliable Quality Monitoring System for Additive Manufacturing Process using Deep Convolutional Neural Networks", 2020, {\it{arXiv:2003.08749.}}

\bibitem{ref50}	A. Saluja, J. Xie, K. Fayazbakhsh, "A closed-loop in-process warping detection system for fused filament fabrication using convolutional neural networks", {\it{J of Manuf Proc,}} 2020, 58, pp. 407-415, doi: 10.1016/j.jmapro.2020.08.036.

\bibitem{ref51}	Z. Jin, Z. Zhang, G.X. Gu, "Automated Real-Time Detection and Prediction of Interlayer Imperfections in Additive Manufacturing Processes Using Artificial Intelligence", {\it{Adv Intell Syst,}} 2019, 2, 1900130, doi: 10.1002/aisy.201900130.

\bibitem{ref52}	V.W.H. Wong, M. Ferguson, K.H. Law, Y.T. Lee, P. Witherell, "Automatic Volumetric Segmentation of Additive Manufacturing Defects with 3D U-Net", in {\it{AAAI 2020 Spring Symposia, Stanford, CA, USA, March, 2020,}} arXiv:2101.08993.

\bibitem{ref53}	V.W.H. Wong, M. Ferguson, K.H. Law, Y.T. Lee, P. Witherell, "Segmentation of Additive Manufacturing Defects Using U-Net," {\it{ASME J Comput Inf Sci Eng,}} 2022, 22(3),  031005, doi: 10.1115/1.4053078.

\bibitem{ref54}	D. Cannizzaro et al., "In-Situ Defect Detection of Metal Additive Manufacturing: An Integrated Framework," in {\it{IEEE Transactions on Emerging Topics in Computing,}} 2022, 10(1), pp. 74-86, doi: 10.1109/TETC.2021.3108844.

\bibitem{ref55}	O. Davtalab, A. Kazemian, X. Yuan, B. Khoshnevis, "Automated inspection in robotic additive manufacturing using deep learning for layer deformation detection," {\it{J Intell Manuf,}} 2022, 33, pp. 771–784, doi: 10.1007/s10845-020-01684-w.

\bibitem{ref56}	J.M. Pearce, A. Petsiuk, "Synthetic-to-real composite semantic segmentation in additive manufacturing." OSF Source file repository. Available: https://osf.io/h8r45. [Accessed Oct. 10, 2022].

\bibitem{ref57}	Thingiverse: an open catalog of computer-aided designs for 3D printing. Available: https://www.thingiverse.com. [Accessed Oct. 10, 2022].

\bibitem{ref58}	MatterControl: 3D Printing Software. Available: https://www.matterhackers.com/store/l/mattercontrol/sk/MKZGTDW6. [Accessed Oct. 10, 2022].

\bibitem{ref59}	H. Löpmeier, "Blender-Gcode-Importer," Available: https://github.com/Heinz-Loepmeier/Blender-Gcode-Import. [Accessed Oct. 10, 2022].

\bibitem{ref60}	Blender: Shader nodes library. Available: \href{https://docs.blender.org/manual/en/3.0/render/shader_nodes/index.html}{https://docs.blender.org}. [Accessed Oct. 10, 2022].


\bibitem{ref61}	Blender: Noise Texture Node. Available: \href{https://docs.blender.org/manual/en/3.0/render/shader_nodes/textures/noise.html}{https://docs.blender.org}. [Accessed Oct. 10, 2022].

\bibitem{ref62}	Blender: Voronoi Texture node. Available: \href{https://docs.blender.org/manual/en/3.0/render/shader_nodes/textures/voronoi.html}{https://docs.blender.org}. [Accessed Oct. 10, 2022].

\bibitem{ref63}	Blender: Principled BSDF. Available: \href{https://docs.blender.org/manual/en/3.0/render/shader\_nodes/shader/principled.html}{https://docs.blender.org}. [Accessed Oct. 10, 2022].

\bibitem{ref64}	Blender: Glossy BSDF. Available: \href{https://docs.blender.org/manual/en/3.0/render/shader\_nodes/shader/glossy.html}{https://docs.blender.org}. [Accessed Oct. 10, 2022].

\bibitem{ref65}	Blender: Diffuse BSDF. Available: \href{https://docs.blender.org/manual/en/3.0/render/shader\_nodes/shader/diffuse.html}{https://docs.blender.org}. [Accessed Oct. 10, 2022].

\bibitem{ref66}	Blender: Transparent BSDF. Available: \href{https://docs.blender.org/manual/en/3.0/render/shader\_nodes/shader/transparent.html}{https://docs.blender.org}. [Accessed Oct. 10, 2022].

\bibitem{ref67}	Blender API. Available: https://docs.blender.org/api/current/. [Accessed Oct. 10, 2022].

\bibitem{ref68}	Blender Compositing. Available: \href{https://docs.blender.org/manual/en/3.0/compositing/introduction.html}{https://docs.blender.org}. [Accessed Oct. 10, 2022].

\bibitem{ref69}	Blender: Material Pass Index. Available: \href{https://docs.blender.org/manual/en/3.0/render/eevee/materials/settings.html}{https://docs.blender.org}. [Accessed Oct. 10, 2022].

\bibitem{ref70}	Blender Cycles. Available: \href{https://docs.blender.org/manual/en/3.0/render/cycles/introduction.html}{https://docs.blender.org}. [Accessed Oct. 10, 2022].

\bibitem{ref71}	O. Ronneberger, P. Fischer, T. Brox, "U-Net: Convolutional Networks for Biomedical Image Segmentation," 2015, {\it{arXiv:1505.04597.}}

\bibitem{ref72}	X. Qin, Z. Zhang, C. Huang, M. Dehghan, O.R. Zaiane, M. Jagersand, "U2-Net: Going Deeper with Nested U-Structure for Salient Object Detection," 2020, {\it{arXiv:2005.09007.}}

\bibitem{ref73}	H. Huang, L. Lin, R. Tong, H. Hu, Q. Zhang, Y. Iwamoto, X. Han, Y.-W. Chen, J. Wu, "UNet 3+: A Full-Scale Connected UNet for Medical Image Segmentation," 2020, {\it{arXiv:2004.08790.}}

\bibitem{ref74}	M. Toldo, U. Michieli, P. Zanuttigh, "Unsupervised Domain Adaptation in Semantic Segmentation via Orthogonal and Clustered Embeddings," 2020, {\it{arXiv:2011.12616.}}

\bibitem{ref75}	J. Yang, C. Li, W. An, H. Ma, Y. Guo, Y. Rong, P. Zhao, J. Huang, "Exploring Robustness of Unsupervised Domain Adaptation in Semantic Segmentation," 2021, {\it{arXiv:2105.10843.}}

\bibitem{ref76}	X. Guo, C. Yang, B. Li, Y. Yuan, "MetaCorrection: Domain-aware Meta Loss Correction for Unsupervised Domain Adaptation in Semantic Segmentation," 2021, {\it{arXiv:2103.05254.}}

\bibitem{ref77}	M. Buda, "U-Net for brain segmentation," 2019. Available: https://pytorch.org/hub/mateuszbuda\_brain-segmentation-pytorch\_unet. [Accessed Oct. 10, 2022].

\bibitem{ref78}	F. Battocchio, "U-Net architecture for Multiclass semantic segmentation," 2020. Available: https://github.com/France1/unet-multiclass-pytorch. [Accessed Oct. 10, 2022].

\bibitem{ref79}	G.E. Hinton and S.T. Roweis, “Stochastic Neighbor Embedding,” in {\it{Advances in Neural Information Processing Systems,}} S. Becker, S. Thrun, K. Obermayer (eds). MIT Press, vol. 15, 2002.

\bibitem{ref80}	L.J.P. van der Maaten, G.E. Hinton, "Visualizing High-Dimensional Data Using t-SNE," {\it{Journal of Machine Learning Research,}} 2008, 9, pp. 2579-2605.

\bibitem{ref81}	T. Xu, W. Chen, P. Wang, F. Wang, H. Li, R. Jin, "CDTrans: Cross-domain Transformer for Unsupervised Domain Adaptation," 2022, {\it{arXiv:2109.06165.}}

\bibitem{ref82}	B. Xie, S. Li, M. Li, C.H. Liu, G. Huang, G. Wang, "SePiCo: Semantic-Guided Pixel Contrast for Domain Adaptive Semantic Segmentation," 2022, {\it{arXiv:2204.08808.}}

\newpage

\bibitem{ref83}	L. Hoyer, D. Dai, L. Van Gool, "HRDA: Context-Aware High-Resolution Domain-Adaptive Semantic Segmentation," in {\it{European Conference on Computer Vision (ECCV),}} 2022, arXiv:2204.13132.

\bibitem{ref84}	P. Han, G. Zhao, "A review of edge-based 3D tracking of rigid objects," {\it{Virtual Reality \& Intelligent Hardware,}} 2019, 1(6), pp. 580-596, doi: 10.1016/j.vrih.2019.10.001.

\bibitem{ref85}	B. Wang, F. Zhong, and X. Qin, "Robust edge-based 3D object tracking with direction-based pose validation," {\it{Multimed Tools Appl,}} 2019, 78, pp. 12307–12331, doi: 10.1007/s11042-018-6727-5.

\bibitem{ref86}	OctoPrint: an open source 3D printer controller application. Available: https://octoprint.org. [Accessed Oct. 10, 2022].

\end{thebibliography}
\end{document}